# Dynamical Mode Recognition of Turbulent Flames in a Swirl-stabilized Annular Combustor by a Time-series Learning Approach

Tao Yang[a], Weiming Xu[a], Liangliang Xu[b*], Peng Zhang[a*]

a Department of Mechanical Engineering, City University of Hong Kong, Kowloon Tong, Kowloon, 999077, Hong Kong

b School of Mechanical Engineering, Shanghai Jiao Tong University, Shanghai 200240, China

[*]Corresponding Author Email: xuliangliang@sjtu.edu.cn (L. Xu), penzhang@cityu.edu.hk (P. Zhang).

**Abstract:** Thermoacoustic instability in annular combustors, essential to aero engines and modern gas turbines, can severely impair operational stability and efficiency, accurately recognizing and understanding various combustion modes is the prerequisite for understanding and controlling combustion instabilities. However, the high-dimensional spatial-temporal dynamics of turbulent flames typically pose considerable challenges to mode recognition. Based on the bidirectional temporal and nonlinear dimensionality reduction models, this study introduces a two-layer bidirectional long short-term memory variational autoencoder (Bi-LSTM-VAE) model to effectively recognize dynamical modes in annular combustion systems. Specifically, leveraging 16 pressure signals from a swirl-stabilized annular combustor, the model maps complex dynamics into a low-dimensional latent space while preserving temporal dependency and nonlinear behavior features through the recurrent neural network structure. The results show that the novel Bi-LSTM-VAE method enables a clear representation of combustion states in two-dimensional state space. Analysis of latent variable distributions reveals distinct patterns corresponding to a wide range of equivalence ratios $\varphi$ and premixed fuel/air mass flow rates Q (SLM), offering novel insights into mode classification and transitions, highlighting this model's potential for deciphering complex thermoacoustic phenomena.

**Introduction**

Annular combustors are a type of combustion chamber widely used in gas turbine engines for aircraft propulsion and power generation. As a critical component in modern gas turbines, annular combustion systems [1] offer advantages in efficiency, emissions, and space utilization. Typically designed as lean premixed combustion systems, they aim to reduce peak temperatures and, consequently, the production of NOx. However, this design choice increases the likelihood of thermoacoustic instabilities in the combustion system [2]. These instabilities are complex and difficult to model so that they present some design and manufacturing challenges. In recent years, many studies have been carried out to investigate thermoacoustic instability [3, 4], where the interaction between heat release and pressure oscillations leads to self-excited sound waves that can significantly affect combustion stability and performance [5].

In practice, identifying combustion mode or flame state is a critical aspect of investigating combustion instability. Understanding these modes is essential for optimizing combustor design and performance. Developing a powerful new tool for the mode discovery and mode recognition of combustion dynamical systems from abundant experiment data of annular combustor is vital. Recently, machine learning has often been used in efforts to study flame stability and dynamical modes. The variational autoencoder (VAE) [6], a probabilistic extension of the traditional autoencoder (AE), is particularly effective for developing nonlinear reduced-order models. Specifically, Omata and Shirayama [7] reported that the convolutional autoencoder is capable of reducing the dimensionality of data and representing the unsteady flow structures. Previous studies have demonstrated that VAE has great potential in reducing data dimensionality and extracting meaningful features.

However, flame stability modes in complex combustion systems are closely tied to temporal correlations, the accurate identification of dynamical modes necessitates a comprehensive understanding of the evolving temporal characteristics. Therefore, Long Short-Term Memory (LSTM) [8] networks have a prominent ability to effectively capture sequential patterns and long-range dependencies in time series data, enabling comprehensive analysis of temporal characteristics within time series datasets [9]. Recently, Lei et al. [10] reported that the LSTM model has better accuracy and robustness for industrial temperature prediction.

In this study, we will develop a novel deep learning approach by integrating LSTM with VAE to study turbulent flames in a swirl-stabilized annular combustor. The dataset includes time series pressure signals from the combustor. The LSTM-VAE model is designed to reduce the dimension of the annular flames systems, enabling comprehensive analysis of their spatial-temporal characteristics.

**Methods / Experimental**
*Experimental Setup and Data Collection*
The data sets for this study were taken from experiments conducted using the annular combustion chamber from Shanghai Jiao Tong University. As shown in Fig. 1(a), a premixed fuel-air mixture is fed into a cylindrical plenum, dividing the flow into 16 swirl-stabilized burners (the zoom figure), equally being spaced around the circumference of the annular combustor.

Each burner has identical configuration, which are comprised of 20 mm diameter swirler with 1.2 swirl number, a bluff body of 26 mm height and 14 mm diameter. The inner and outer walls of the chamber are 135 mm and 435 mm long respectively, with diameters of 182 mm and 280 mm. The fuel is methane and mixed with air through the inlets. The mass flow rate is set to a value of 1600, 2240, 2560, and 2880 SLM (standard liter per minute) and the equivalence ratio, $\varphi$ is varied in the range of 0.65 to 0.95. 16 dynamic pressure sensors are mounted at eight azimuthal and two longitudinal positions flush with the inner wall of the injector tubes and recorded with a sampling frequency of 20kHz.

Annular combustion was performed and measured under different equivalence ratio $\varphi$ and premixed fuel/air mass flow rate $Q$ (SLM) so that total 23 data sets were generated. Each data set contains 16 time-series signals of pressure fluctuation within 2.0 s, as shown in Fig. 1(b).

*Nonlinear Latent Representations via Deep Probabilistic Neural Network*
Variational autoencoder (VAE), a variant of the autoencoder (AE) that combines variational inference to learn a latent space representation of data, is extensively utilized as a potent dimensionality reduction tool in pattern analysis and combustion instability prediction. Our recent work [11] utilized this neural network for dynamical mode recognition of various flame oscillator systems.

The conventional VAE fundamentally consists of three parts: an encoder, a reparameterization process, and a decoder. The temporal nature of the data dictates the use of a long short-term memory (LSTM) neural network for the encoder and decoder, as LSTM layers are particularly adept at capturing the complex, time-dependent patterns inherent in flame dynamics. As shown in Fig. 2, we introduce a two-layer bidirectional LSTM architecture in this study. The bidirectional structure processes data in both forward and backward directions, allowing the network to incorporate both past and future temporal contexts. By integrating VAE and Bi-LSTM, we propose a dual-layer bidirectional long short-term memory variational autoencoder (Bi-LSTM-VAE) to seamlessly integrate sequential modeling and latent space learning to recognize and predict complex dynamical modes in annular combustion systems. More details on the proposed Bi-LSTM-VAE model can be found in our recent work [12].

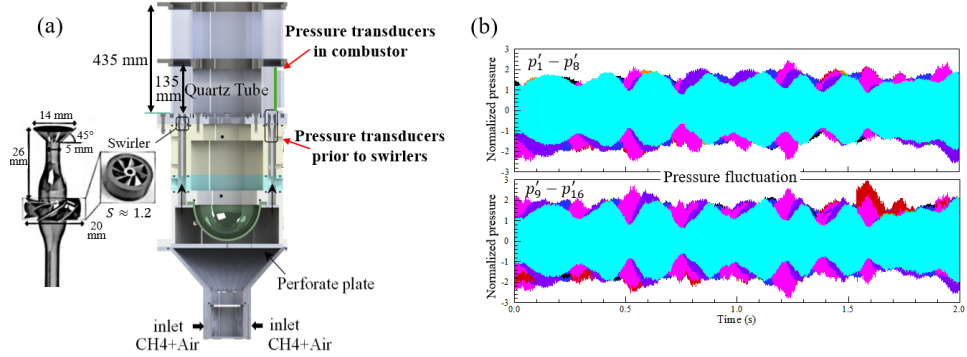

Fig. 1 (a) Schematic diagram of experiments with combustor and measurement devices; (b) Pressure fluctuation of sixteen transducers.

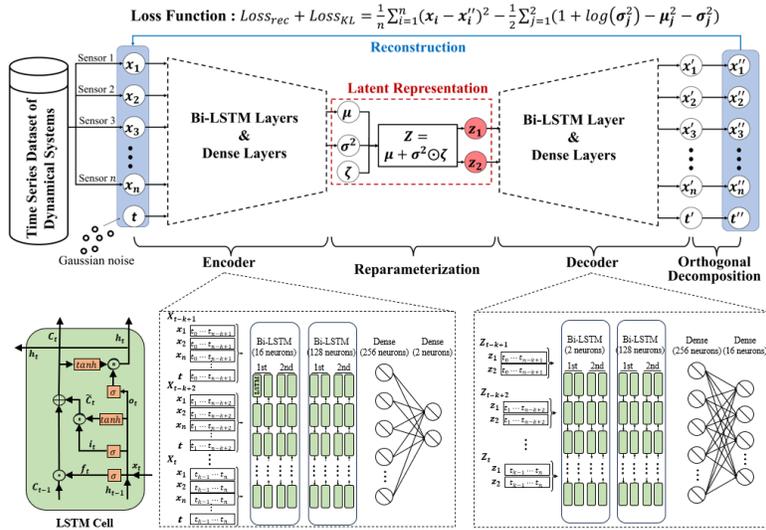

Fig. 2. The Bi-LSTM-VAE framework for temporal dynamics and latent space representation consists of four modules: Encoder, Reparameterization, Decoder, and Orthogonal Decomposition.

The proposed model is trained through the minimization of mean squared error (MSE) and Kullback-Leibler (KL) divergence between input and reconstructed data. The Adam optimization algorithm is selected as the optimizer for the model. The training concludes either after 100 calculations of no improvement in validation set loss or upon reaching the pre-defined epoch limit of 5000. The models and training pipeline were implemented on NVIDIA A800 GPUs for about 6 hours by using the Torch 1.12 deep learning framework.

**Results and Discussion**

Figure 3 shows the results on the identification of three flame modes via the Bi-LSTM-VAE framework. With varying $Q$ and $\varphi$, the 23 cases have their phase distributions in the two-dimensional latent space of $Z_1$ and $Z_2$, which can manifest flame dynamics. As a result, we found that theses phase distributions totally exhibit distinct three patterns:

Mode I, single-variable-dependent pattern in Fig. 3(b), where the phase trajectory only changes in the $Z_1$ direction and the $Z_2$ is almost a constant, while it may be different in different cases;

Mode II, double-variable-dependent pattern in Fig. 3(c) in which the phase trajectory both changes in the $Z_1$ and $Z_2$ directions;

Mode III, dual-stable pattern in Fig. 3(d-f), where the phase points scatter around the two limit points.

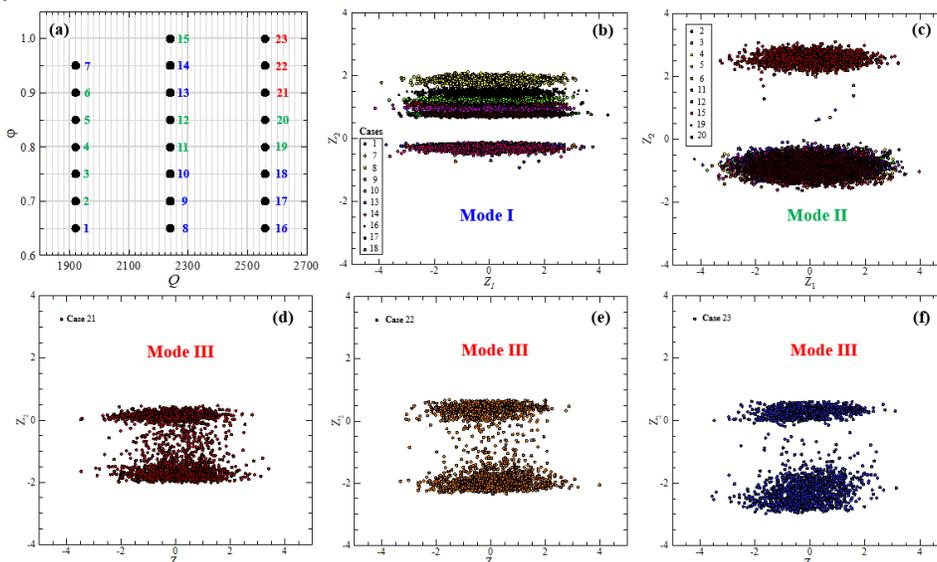

Fig. 3. (a) Distinct flame modes in the space of $Q$ and $\varphi$ for all cases. (b) Phase distributions in the latent space of various modes: (b) Mode I, (c) Mode II, and (d-f) Mode III.

In Fig. 3(a), the three dynamical modes are labeled in the space of $Q$ and $\varphi$ for all cases and there are some observations: first, most of Mode I cases occurs in the regime with a lower $\varphi$ or a higher $\varphi$; second, the Mode II cases appear in the regime of middle $\varphi$ and the range of $\varphi$ gradually narrows with increasing $Q$; third, the Mode III cases locates at the regime of high $Q$ and $\varphi$. In detail, we compared the phase distributions of three modes in time domain, taking three representative cases, as shown in Fig. S1-S3. All flames exhibit periodical oscillations, but in distinct manners. The flames in Mode I show a small amplitude fluctuation, while those in Mode II are unstable obviously. Interestingly, these Mode III cases present another instability mode.

**Conclusions**

An innovative deep neural network of Bi-LSTM-VAE is established to map a 16-dimenisonal pressure time series data into a 2-dimensional latent space for classifying the phase distributions, which can manifest flame dynamics to some extent. The datasets of pressure time series collected from the turbulent flames in a swirl-stabilized annular combustor were utilized to classify three distinct dynamical modes with different patterns of phase distributions.

The present results show that the proposed approach is effective in identifying dynamical modes of complex combustion systems through comparing, analyzing, and classifying the distributions of latent variables, resulting from the dimensionality reduction model, which could consist of a low dimensional state space for the dynamical processes in physical space.

**Acknowledgements**

This work is supported by the National Natural Science Foundation of China (No. 52176134) and partially by the APRC-CityU New Research Initiatives/Infrastructure Support from Central of City University of Hong Kong (No. 9610601).